\documentclass[runningheads]{llncs}
\usepackage{graphicx, lineno, hyperref, amsmath, booktabs, amssymb, array, multirow}

\begin{document}
\title{Detection of Anomalies in Large-Scale Accounting Data using Deep Autoencoder Networks}
%
%
\author{Marco Schreyer\inst{1} \and
Timur Sattarov\inst{2} \and Damian Borth\inst{1} \and \\ Andreas Dengel\inst{1} \and Bernd Reimer\inst{2}}
\authorrunning{Schreyer, Sattarov et al.}
%
\institute{
German Research Center for Artificial Intelligence (DFKI) GmbH \\
Trippstadter Stra{\ss}e 112, Kaiserslautern, Germany \\
\email{firstname.lastname@dfki.de} \\ \and
PricewaterhouseCoopers (PwC) GmbH, \\
Friedrichstra{\ss}e 14, Stuttgart, Germany \\
\email{lastname.firstname@pwc.com}
}
\maketitle              

\begin{abstract}
Learning to detect fraud in large-scale accounting data is one of the long-standing challenges in financial statement audits or fraud investigations. Nowadays, the majority of applied techniques refer to handcrafted rules derived from known fraud scenarios. While fairly successful, these rules exhibit the drawback that they often fail to generalize beyond known fraud scenarios and fraudsters gradually find ways to circumvent them. To overcome this disadvantage and inspired by the recent success of deep learning we propose the application of deep autoencoder neural networks to detect anomalous journal entries. We demonstrate that the trained network's reconstruction error obtainable for a journal entry and regularized by the entry's individual attribute probabilities can be interpreted as a highly adaptive anomaly assessment. Experiments on two real-world datasets of journal entries, show the effectiveness of the approach resulting in high f\textsubscript{1}-scores of 32.93 (dataset A) and 16.95 (dataset B) and less false positive alerts compared to state of the art baseline methods. Initial feedback received by chartered accountants and fraud examiners underpinned the quality of the approach in capturing highly relevant accounting anomalies.

\keywords{Accounting Information Systems \and Computer Assisted Audit Techniques (CAATs) \and Journal Entry Testing \and Forensic Accounting \and Fraud Detection \and Forensic Data Analytics \and Deep Learning}

\end{abstract}

\section{Motivation}
\label{sec:motivation}

The Association of Certified Fraud Examiners estimates in its Global Fraud Study 2016 \cite{ACFE2016} that the typical organization lost 5\% of its annual revenues due to fraud. The term "fraud" refers to "the abuse of one's occupation for personal enrichment through the deliberate misuse of an organization's resources or assets" \cite{Wells2017}. A similar study, conducted by PwC, revealed that nearly a quarter (22\%) of respondents experienced losses between \$100'000 and \$1 million due to fraud \cite{PWC2016}. The study also showed that financial statement fraud caused by far the highest median loss of the surveyed fraud schemes\footnote{The ACFE study encompasses an analysis of 2'410 cases of occupational fraud investigated between January 2014 and October 2015 that occurred in 114 countries. The PwC study encompasses over 6'000 correspondents that experienced economic crime in the last 24 months.}. 

\begin{figure}[t]
	\hspace*{0.0cm} \includegraphics[scale=0.50, angle=0, trim={0.6cm 2.8cm 0.0 0.0}]{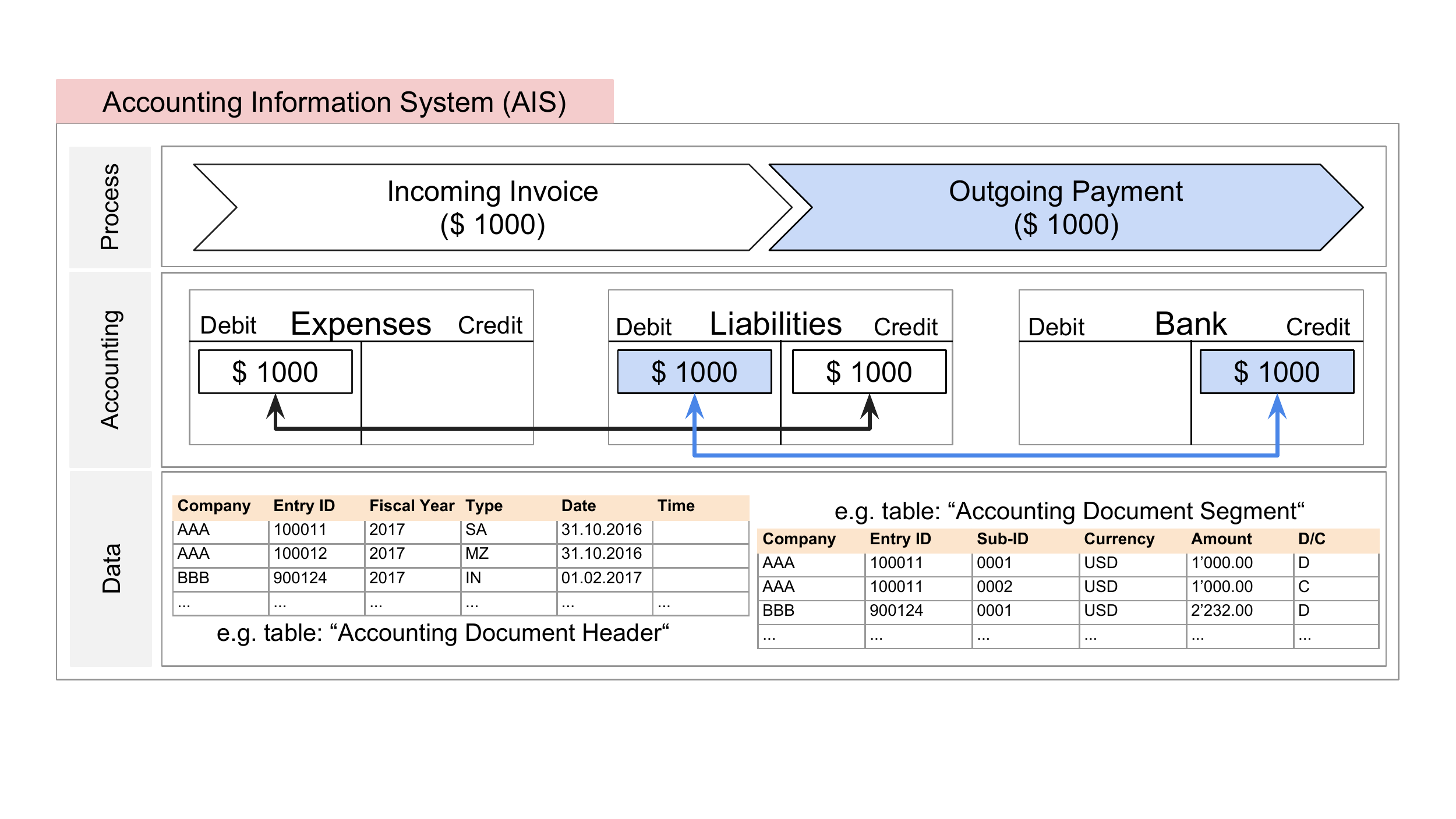}
	\caption{Hierarchical view of an Accounting Information System (AIS) that records distinct layer of abstractions, namely (1) the business process, (2) the accounting and (3) technical journal entry information in designated database tables.}
	\label{fig:ais_system}
\end{figure}

At the same time, organizations accelerate the digitization and reconfiguration of business processes \cite{McKinsey2014} affecting in particular Accounting Information Systems (AIS) or more generally Enterprise Resource Planning (ERP) systems. Steadily, these systems collect vast quantities of electronic evidence at an almost "atomic" level. This holds in particular for the journal entries of an organization recorded in its general ledger and sub-ledger accounts. SAP, one of the most prominent enterprise software providers, estimates that approx. 76\% of the world's transaction revenue touches one of their ERP systems \cite{SAP2017}. Figure \ref{fig:ais_system} depicts a hierarchical view of an AIS recording process of journal entry information in designated database tables.

To detect potentially fraudulent activities international audit standards require the direct assessment of journal entries \cite{AICPA2002},\cite{IFAC2009}. Nowadays, the majority of applied techniques to examine journal entries refer to rules defined by experienced chartered accountants or fraud examiners that are handcrafted and often executed manually. The techniques, usually based on known fraud scenarios, are often referred to as "red-flag" tests (e.g. postings late at night, multiple vendor bank account changes, backdated expense account adjustments) or statistical analyses (e.g. Benford's Law \cite{Benford1938}, time series evaluation). Unfortunately, they often don't generalize beyond historical fraud cases already known and therefore fail to detect novel schemes of fraud. In addition, such rules become rapidly outdated while fraudsters adaptively find ways to circumvent them. 

Recent advances in deep learning \cite{LeCun2015} enabled scientists to extract complex nonlinear features from raw sensory data leading to breakthroughs across many domains e.g. computer vision \cite{Krizhevsky2012} and speech recognition \cite{Mikolov2013a}. Inspired by those developments we propose the application of deep autoencoder neural networks to detect anomalous journal entries in large volumes of accounting data. We envision this automated and deep learning based examination of journal entries as an important supplement to the accountants and forensic examiners toolbox \cite{Pedrosa2014}.


In order to conduct fraud, perpetrators need to deviate from regular system usage or posting pattern. Such deviations are recorded by a very limited number of "anomalous" journal entries and their respective attribute values. Based on this observation we propose a novel scoring methodology to detect anomalous journal entries in large scale accounting data. The scoring considers (1) the magnitude of a journal entry's reconstruction error obtained by a trained deep autoencoder network and (2) regularizes it by the entry's individual attribute probabilities. This anomaly assessment is highly adaptive to the often varying attribute value probability distributions of journal entries. Furthermore, it allows to flag entries as "anomalous" if they exceed a predefined scoring threshold. We evaluate the proposed method based on two anonymized real-world datasets of journal entries extracted from large-scale SAP ERP systems. The effectiveness of the proposed method is underpinned by a comparative evaluation against state of the art anomaly detection algorithms.

In section \ref{sec:relatedwork} we provide an overview of the related work. Section \ref{sec:methodology} follows with a description of the autoencoder network architecture and presents the proposed methodology to detect accounting anomalies. The experimental setup and results are outlined in section \ref{sec:experiments} and section \ref{sec:results}. In section \ref{sec:conclusion} the paper concludes with a summary of the current work and future directions of research.

\section{Related work}
\label{sec:relatedwork}

The task of detecting fraud and accounting anomalies has been studied both by practitioners \cite{Wells2017} and academia \cite{Amani2017}. Several references describe different fraud schemes and ways to detect unusual and "creative" accounting practices \cite{Singleton2006}. The literature survey presented hereafter focuses on (1) the detection of fraudulent activities in Enterprise Resource Planning (ERP) data and (2) the detection of anomalies using autoencoder networks.

\subsection{Fraud Detection in Enterprise Resource Planning (ERP) Data}

The forensic analysis of journal entries emerged with the advent of Enterprise Resource Planning (ERP) systems and the increased volume of data recorded by such systems. Bay et al. in \cite{Bay2002} used Naive Bayes methods to identify suspicious general ledger accounts, by evaluating attributes derived from journal entries measuring any unusual general ledger account activity. Their approach was enhanced by McGlohon et al. applying link analysis to identify (sub-) groups of high-risk general ledger accounts \cite{McGlohon2009}. 

Kahn et al. in \cite{Khan2009} and \cite{Khan2010} created transaction profiles of SAP ERP users. The profiles are derived from journal entry based user activity pattern recorded in two SAP R/3 ERP system in order to detect suspicious user behavior and segregation of duties violations. Similarly, Islam et al. used SAP R/3 system audit logs to detect known fraud scenarios and collusion fraud via a "red-flag" based matching of fraud scenarios \cite{Islam2010}. 

Debreceny and Gray in \cite{Debreceny2010} analyzed dollar amounts of journal entries obtained from 29 US organizations. In their work, they searched for violations of Benford's Law \cite{Benford1938}, anomalous digit combinations as well as unusual temporal pattern such as end-of-year postings. More recently, Poh-Sun et al. in \cite{Seow2016} demonstrated the generalization of the approach by applying it to journal entries obtained from 12 non-US organizations.

Jans et al. in \cite{Jans2010} used latent class clustering to conduct an uni- and multivariate clustering of SAP ERP purchase order transactions. Transactions significantly deviating from the cluster centroids are flagged as anomalous and are proposed for a detailed review by auditors. The approach was enhanced in \cite{Jans2011} by a means of process mining to detect deviating process flows in an organization procure to pay process.

Argyrou et al. in \cite{Argyrou2012} evaluated self-organizing maps to identify "suspicious" journal entries of a shipping company. In their work, they calculated the Euclidean distance of a journal entry and the code-vector of a self-organizing maps best matching unit. In subsequent work, they estimated optimal sampling thresholds of journal entry attributes derived from extreme value theory \cite{Argyrou2013}.

Concluding from the reviewed literature, the majority of references draw either (1) on historical accounting and forensic knowledge about various "red-flags" and fraud schemes or (2) on traditional non-deep learning techniques. As a result and in agreement with \cite{Wang2010}, we see a demand for unsupervised and novel approaches capable to detect so far unknown scenarios of fraudulent journal entries.

\subsection{Anomaly Detection using Autoencoder Neural Networks}

Nowadays, autoencoder networks have been widely used in image classification \cite{Hinton2006}, machine translation \cite{Lauly2014} and speech processing \cite{Toth2004} for their unsupervised data compression capabilities. To the best of our knowledge Hawkins et al. and Williams et al. were the first who proposed autoencoder networks for anomaly detection \cite{Hawkins2002}, \cite{Williams2002}. 

Since then the ability of autoencoder networks to detect anomalous records was demonstrated in different domains such as X-ray images of freight containers \cite{Andrews2016}, the KDD99, MNIST, CIFAR-10 as well as several other datasets obtained from the UCI Machine Learning Repository\footnote{https://archive.ics.uci.edu/ml/datasets.html} \cite{Dau2014}, \cite{An2016}, \cite{Zhai2016}. In \cite{Zhou2017} Zhou and Paffenroth enhanced the standard autoencoder architecture by an additional filter layer and regularization penalty to detect anomalies.

More recently, autoencoder networks have been also applied in the domain of forensic data analysis. Cozzolino and Verdoliva used the autoencoder reconstruction error to detect pixel manipulations of images \cite{Cozzolino2017}. In \cite{Davino2017} the method was enhanced by recurrent neural networks to detect forged video sequences. Lately, Paula et al. in \cite{Paula2017} used autoencoder networks in export controls to detect traces of money laundry and fraud by analyzing volumes of exported goods.

To the best of our knowledge, this work presents the first deep learning inspired approach to detect anomalous journal entries in real-world and large-scale accounting data. 


\section{Detection of Accounting Anomalies}
\label{sec:methodology}

In this section we introduce the main elements of autoencoder neural networks. We furthermore describe how the reconstruction error of such networks can be used to detect anomalous journal entries in large-scale accounting data.

\subsection{Deep Autoencoder Neural Networks}

\begin{figure}[t]
	\centering
	\includegraphics[scale=0.5, angle=90, trim= 200 0 200 80]{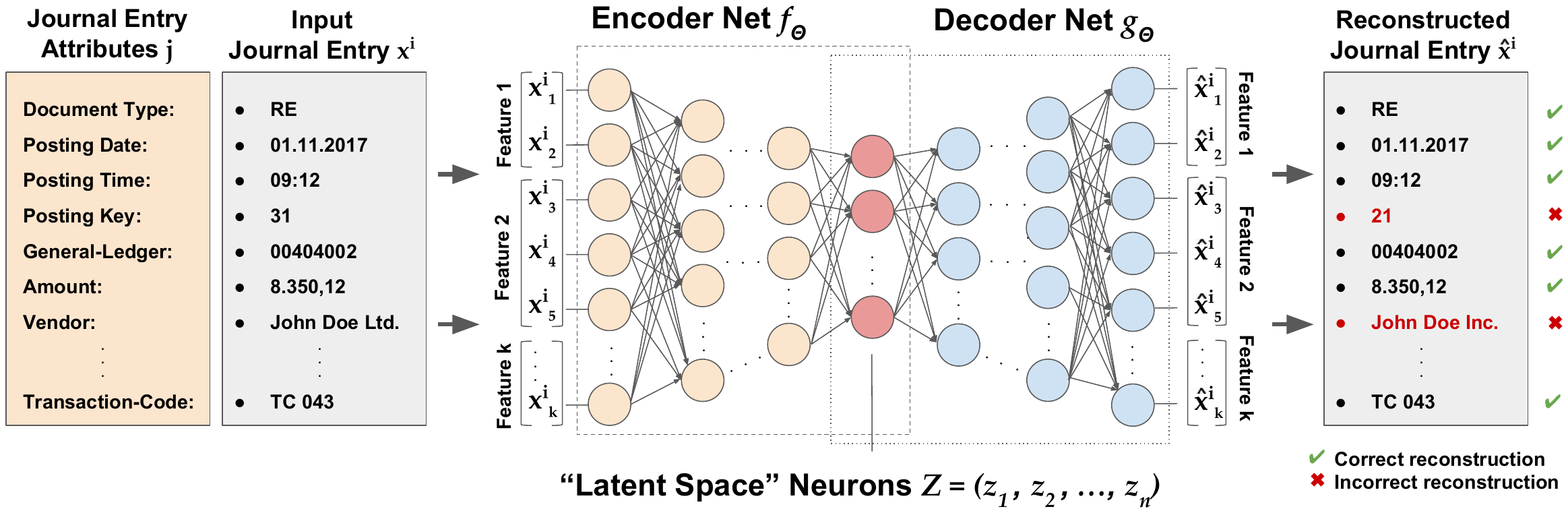}
	\caption{Schematic view of an autoencoder network comprised of two non-linear mappings (fully connected feed forward neural networks) referred to as encoder $f_\theta:\mathbb{R}^{d_x}\mapsto \mathbb{R}^{d_z}$ and decoder $g_\theta:\mathbb{R}^{d_z}\mapsto \mathbb{R}^{d_y}$.}
	\label{fig:architecture}
\end{figure}

We consider the task of training an autoencoder neural network using a set of $N$ journal entries $X = \{x^{1}, x^{2}, ..., x^{n}\}$ where each journal entry $x^{i}$ consists of a tuple of $K$ attributes $x^{i} = (x_{1}^{i}, x_{2}^{i}, ... , x_{j}^{i}, ... ,x_{k}^{i})$. Thereby, $x_{j}^{i}$ denotes the $j^{th}$ attribute of the $i^{th}$ journal entry. The individual attributes $x_{j}$ encompass a journal entry's accounting specific details e.g. posting type, posting date, amount, general-ledger. Furthermore, $n_{j}^{i}$ counts the occurrence of a particular attribute value of attribute $x_{j}$ e.g. a specific document type or account. 

An autoencoder or replicator neural network defines a special type of feed-forward multilayer neural network that can be trained to reconstruct its input. The difference between the original input and its reconstruction is referred to as reconstruction error. Figure \ref{fig:architecture} illustrates a schematic view of an autoencoder neural network. In general, autoencoder networks are comprised of two nonlinear mappings referred to as encoder $f_\theta$ and decoder $g_\theta$ network \cite{Rumelhart1986}. Most commonly the encoder and the decoder are of symmetrical architecture consisting of several layers of neurons each followed by a nonlinear function and shared parameters $\theta$. The encoder mapping $f_\theta(\cdot)$ maps an input vector $x^{i}$ to a compressed representation $z^{i}$ in the latent space $Z$. This latent representation $z^{i}$ is then mapped back by the decoder $g_\theta(\cdot)$ to a reconstructed vector $\hat{x}^{i}$ of the original input space. Formally, the non-linear encoder and decoder mapping of an autoencoder encompassing several layers of neurons can be defined by:
\begin{equation}
	f^l_\theta(\cdot) = \sigma^{l}(W^{l}(f^{l-1}_\theta(\cdot))+b^{l}), \text{ and } g^l_\theta(\cdot) = \sigma'^{l}(W'^{l}(g^{l-1}_\theta(\cdot))+d^{l}),
\end{equation}

\noindent where $\sigma$ and $\sigma'$ denote non-linear activations e.g. the sigmoid function, $\theta$ denote the model parameters $\{W, b, W', d\}$, $W \in \mathbb{R}^{d_x \times d_z}, W' \in \mathbb{R}^{d_z \times d_y}$ are weight matrices, $b \in \mathbb{R}^{d_z}, d\in \mathbb{R}^{d_y}$ are offset bias vectors and $l$ denotes the number of hidden layers. 

In an attempt to achieve $x^{i} \approx \hat{x}^{i}$ the autoencoder is trained to learn a set of optimal encoder-decoder model parameters $\theta^*$ that minimize the dissimilarity of a given journal entry $x^{i}$ and its reconstruction $\hat{x}^{i} = g_\theta(f_\theta(x^{i}))$ as faithfully as possible. Thereby, the autoencoder training objective is to learn a model that optimizes:
\begin{equation}
	\arg\min_{\theta} \|X - g_\theta(f_\theta(X))\|,
\end{equation}

\noindent for all journal entries $X$. As part of the network training one typically minimizes a loss function $\mathcal{L_{\theta}}$ defined by the squared reconstruction loss or, as used in our experiments, the cross-entropy loss given by:
\begin{equation}
	\mathcal{L_{\theta}}(x^{i};\hat{x}^{i}) = \frac{1}{n}\sum_{i=1}^{n}\sum_{j=1}^{k} x^{i}_{j} ln(\hat{x}^{i}_{j}) + (1-x^{i}_{j}) ln(1-\hat{x}^{i}_{j}),
\end{equation}

\noindent for a set of $n$-journal entries $x^{i}$, $i=1,...,n$ and their respective reconstructions $\hat{x}^{i}$ over all journal entry attributes $j=1,...,k$. For binary encoded attribute values, as used in this work, the $\mathcal{L_{\theta}}(x^{i}; \hat{x}^{i})$ measures the deviation between two independent multivariate Bernoulli distributions, with mean $x$ and mean $\hat{x}$ respectively \cite{Bengio2013}.

To prevent the autoencoder from learning the identity function the number of neurons of the networks hidden layers are reduced indicating $\mathbb{R}^{d_x} > \mathbb{R}^{d_z}$ (usually referred to as "bottleneck" architecture). Imposing such a constraint onto the network's hidden layer forces the autoencoder to learn an optimal set of parameters $\theta^*$ that result in a "compressed" model of the most prevalent journal entry attribute value distributions and their dependencies. 


\subsection{Classification of Accounting Anomalies}
\label{sec:anomalyclassification}

To detect anomalous journal entries we first have to define "normality" with respect to accounting data. We assume that the majority of journal entries recorded within an organizations' ERP system relate to regular day-to-day business activities. In order to conduct fraud, perpetrators need to deviate from the "normal". Such deviating behavior will be recorded by a very limited number of journal entries and their respective attribute values. We refer to journal entries exhibiting such deviating attribute values as \textit{accounting anomalies}.

When conducting a detailed examination of real-world journal entries, recorded in large-scaled ERP systems, two prevalent characteristics can be observed: First, journal entry attributes exhibit a high variety of distinct attribute values and second, journal entries exhibit strong dependencies between certain attribute values e.g. a document type that is usually posted in combination with a certain general ledger account. Derived from this observation and similarly to Breunig et al. in \cite{Breunig2000} we distinguish two classes of anomalous journal entries, namely \textit{global} and \textit{local anomalies}:

\textbf{Global accounting anomalies}, are journal entries that exhibit unusual or rare individual attribute values. Such anomalies usually relate to skewed attributes e.g. rarely used ledgers, or unusual posting times. Traditionally, "red-flag" tests performed by auditors during an annual audit, are designed to capture this type of anomaly. However, such tests often result in a high volume of false positive alerts due to events such as reverse postings, provisions and year-end adjustments usually associated with a low fraud risk. Furthermore, when consulting with auditors and forensic accountants, "global" anomalies often refer to "error" rather than "fraud". 

\textbf{Local accounting anomalies}, are journal entries that exhibit an unusual or rare combination of attribute values while their individual attribute values occur quite frequently e.g. unusual accounting records, irregular combinations of general ledger accounts, user accounts used by several accounting departments. This type of anomaly is significantly more difficult to detect since perpetrators intend to disguise their activities by imitating a regular activity pattern. As a result, such anomalies usually pose a high fraud risk since they correspond to processes and activities that might not be conducted in compliance with organizational standards.

In regular audits, accountants and forensic examiners desire to detect journal entries corresponding to both anomaly classes that are "suspicious" enough for a detailed examination. In this work, we interpret this concept as the detection of (1) any unusual individual attribute value or (2) any unusual combination of attribute values observed. This interpretation is also inspired by earlier work of Das and Schneider \cite{DAS2007} on the detection of anomalous records in categorical datasets.


\subsection{Scoring of Accounting Anomalies}
\label{sec:anomalyscoring}

Based on this interpretation we propose a novel anomaly score to detect global and local anomalies in real-world accounting datasets. Our score accounts for both of the observed characteristics, namely (1) any "unusual" \textit{attribute value occurrence} (global anomaly) and (2) any "unusual" \textit{attribute value co-occurrence} (local anomaly):



\textbf{Attribute value occurrence}: To account for the observation of unusual or rare attribute values we determine for each value $x_{j}$ its probability of occurrence in the population of journal entries. This can be formally defined by $\frac{n_{j}^{i}}{N}$ were $N$ counts the total number of journal entries. For example, the probability of observing a specific general ledger or posting key in $X$. In addition, we obtain the sum of individual attribute value log-probabilities $P(x^{i})=\sum_{j=1}^{k} \ln (1 + \frac{n^{i}_{j}}{N})$ for each journal entry $x^{i}$ over all its $j$ attributes. Finally, we obtain a min-max normalized attribute value probability score $AP$ denoted by:
\begin{equation}
	AP(x^{i}) = \frac{P(x^{i}) - P_{\min}}{P_{\max} - P_{min}},
\end{equation}

\noindent for a given journal entry $x^{i}$ and its individual attributes $x^{i}_{j}$, where $P_{max}$ and $P_{min}$ denotes $\min$- and $\max$-values of the summed individual attribute value log-probabilities given by $P$.

\textbf{Attribute value co-occurrence}: To account for the observation of irregular attribute value co-occurrences and to target local anomalies, we determine a journal entry's reconstruction error derived by training a deep autoencoder neural network. For example, the probability of observing a certain general ledger account in combination with a specific posting type within the population of all journal entries $X$. Anomalous co-occurrences are hardly learned by the network and can therefore not be effectively reconstructed from their low-dimensional latent representation. Therefore, such journal entries will result in a high reconstruction error. Formally, we derive the trained autoencoder network's reconstruction error $E$ as the squared- or $L2$-difference $E_{\theta^*}(x^{i};\hat{x}^{i}) = \frac{1}{k} \sum_{j=1}^{k}{(x^{i}_{j} - \hat{x}^{i}_{j})}^2$ for a journal entry $x^i$ and its reconstruction $\hat{x}^i$ under optimal model parameters $\theta^*$. Finally, we calculate the normalized reconstruction error $RE$ denoted by:

\begin{equation}
	RE_{\theta^*}(x^{i};\hat{x}^{i}) = \frac{E_{\theta^*}(x^i;\hat{x}^{i}) - E_{\theta^*, min}}{E_{\theta^*, max} - E_{\theta^*, min}},
\end{equation}

\noindent where $E_{min}$ and $E_{max}$ denotes the min- and max-values of the obtained reconstruction errors given by $E_{\theta^*}$. 

\textbf{Accounting anomaly scoring}: Observing both characteristics for a single journal entry, we can reasonably conclude (1) if an entry is anomalous and (2) if it was created by a "regular" business activity. It also implies that we have seen enough evidence to support our judgment. To detect global and local accounting anomalies in real-world audit scenarios we propose to score each journal entry $x^i$ by its reconstruction error $RE$ regularized by its normalized attribute probabilities $AP$ given by:
\begin{equation}
	AS(x^{i};\hat{x}^{i}) = \alpha \times RE_{\theta^*}(x^{i};\hat{x}^{i}) + (1-\alpha) \times AP(x^{i}),
\end{equation} 

\noindent for each individual journal entry $x^{i}$ and optimal model parameters $\theta^*$. We introduce $\alpha$ as a factor to balance both characteristics. In addition, we flag a journal entry as anomalous if its anomaly score $AS$ exceeds a threshold parameter $\beta$, as defined by:
\begin{equation}
	AS(x^{i};\hat{x}^{i}) = 
    \begin{cases}
    	AS(x^{i};\hat{x}^{i}), & \text{$AS(x^{i};\hat{x}^{i}) \geq \beta$}  \\
        0, & \text{otherwise}
    \end{cases},
\end{equation}  

\noindent for each individual journal entry $x^{i}$ under optimal model parameters $\theta^*$.

\section{Experimental Setup and Network Training}
\label{sec:experiments}

In this section we describe the experimental setup and model training. We evaluated the anomaly detection performance of nine distinct autoencoder architectures based on two real-world datasets of journal entries.

\subsection{Datasets and Data Preparation}

\begin{table}[t]
  \centering
  \fontsize{8}{6}\selectfont
  \begin{tabular}{llll}
    \toprule &AE&Fully Connected Layers and Neurons\\
    \midrule
    &AE 1&[401; 576]-3-[401; 576]\\
    &AE 2&[401; 576]-4-3-4-[401; 576]\\
    &AE 3&[401; 576]-8-4-3-4-8-[401; 576]\\
    &AE 4&[401; 576]-16-8-4-3-4-8-16-[401; 576]\\
    &AE 5&[401; 576]-32-16-8-4-3-4-8-16-32-[401; 576]\\
    &AE 6&[401; 576]-64-32-16-8-4-3-4-8-16-32-64-[401; 576]\\
    &AE 7&[401; 576]-128-64-32-16-8-4-3-4-8-16-32-64-128-[401; 576]\\
    &AE 8&[401; 576]-256-128-64-32-16-8-4-3-4-8-16-32-64-128-256-[401; 576]\\
    &AE 9&[401; 576]-512-256-128-64-32-16-8-4-3-4-8-16-32-64-128-256-512-[401; 576]\\
    \bottomrule
    \\
  \end{tabular}
 \caption{Evaluated architecture ranging from shallow architectures (AE 1) encompassing a single fully connected hidden layer to deep architectures (AE 9) encompassing several hidden layers.}
  \label{fig:topologytable}
 \end{table}

Both datasets have been extracted from SAP ERP instances, denoted SAP ERP dataset A and dataset B in the following, encompassing the entire population of journal entries of a single fiscal year. In compliance with strict data privacy regulations, all journal entry attributes have been anonymized using an irreversible one-way hash function during the data extraction process. To ensure data completeness, the journal entry based general ledger balances were reconciled against the standard SAP trial balance reports e.g. the SAP "RFBILA00" report.

In general, SAP ERP systems record a variety of journal entry attributes predominantly in two tables technically denoted by "BKPF" and "BSEG". The table "BKPF" - "Accounting Document Header" contains the meta information of a journal entry e.g., document id, type, date, time, currency. The table "BSEG" - "Accounting Document Segment", also referred to journal entry line-items, contains the entry details e.g., posting key, general ledger account, debit and credit information, amount. We extracted a subset of 6 (dataset A) 
and 10 (dataset B) 
most discriminative attributes of the "BKPF" and "BSEG" tables. 

The majority of attributes recorded in ERP systems correspond to categorical (discrete) variables, e.g. posting date, account, posting type, currency. In order to train autoencoder neural networks, we preprocessed the categorical journal entry attributes to obtain a binary ("one-hot" encoded) representation of each journal entry. This preprocessing resulted in a total of 401 encoded dimensions for dataset A and 576 encoded dimensions for dataset B.

To allow for a detailed analysis and quantitative evaluation of the experiments we injected a small fraction of synthetic \textit{global} and \textit{local anomalies} into both datasets. Similar to real audit scenarios this resulted in highly unbalanced class distribution of "anomalous" vs. "regular" day-to-day entries. The injected global anomalies are comprised of attribute values not evident in the original data while the local anomalies exhibit combinations of attribute value subsets not occurring in the original data. The true  labels ("ground truth") are available for both datasets. Each journal entry is labeled as either synthetic \textit{global anomaly}, synthetic \textit{local anomaly} or non-synthetic \textit{regular entry}. The following descriptive statistics summarize both datasets:

\begin{itemize}
\setlength{\itemindent}{-.2in}

\item \begin{flushleft}\textbf{Dataset A} contains a total of 307'457 journal entry line items comprised of 6 categorical attributes. In total 95 (0.03\%) synthetic anomalous journal entries have been injected into dataset. These entries encompass 55 (0.016\%) global anomalies and 40 (0.015\%) local anomalies.\end{flushleft}

\item \begin{flushleft}\textbf{Dataset B} contains a total of 172'990 journal entry line items comprised of 10 categorical attributes. In total 100 (0.06\%) synthetic anomalous journal entries have been injected into the dataset. These entries encompass 50 (0.03\%) global anomalies and 50 (0.03\%) local anomalies.\end{flushleft}

\end{itemize}

\subsection{Autoencoder Neural Network Training}

\begin{figure}[t]
\minipage{0.5\textwidth}
	\includegraphics[width=\linewidth]{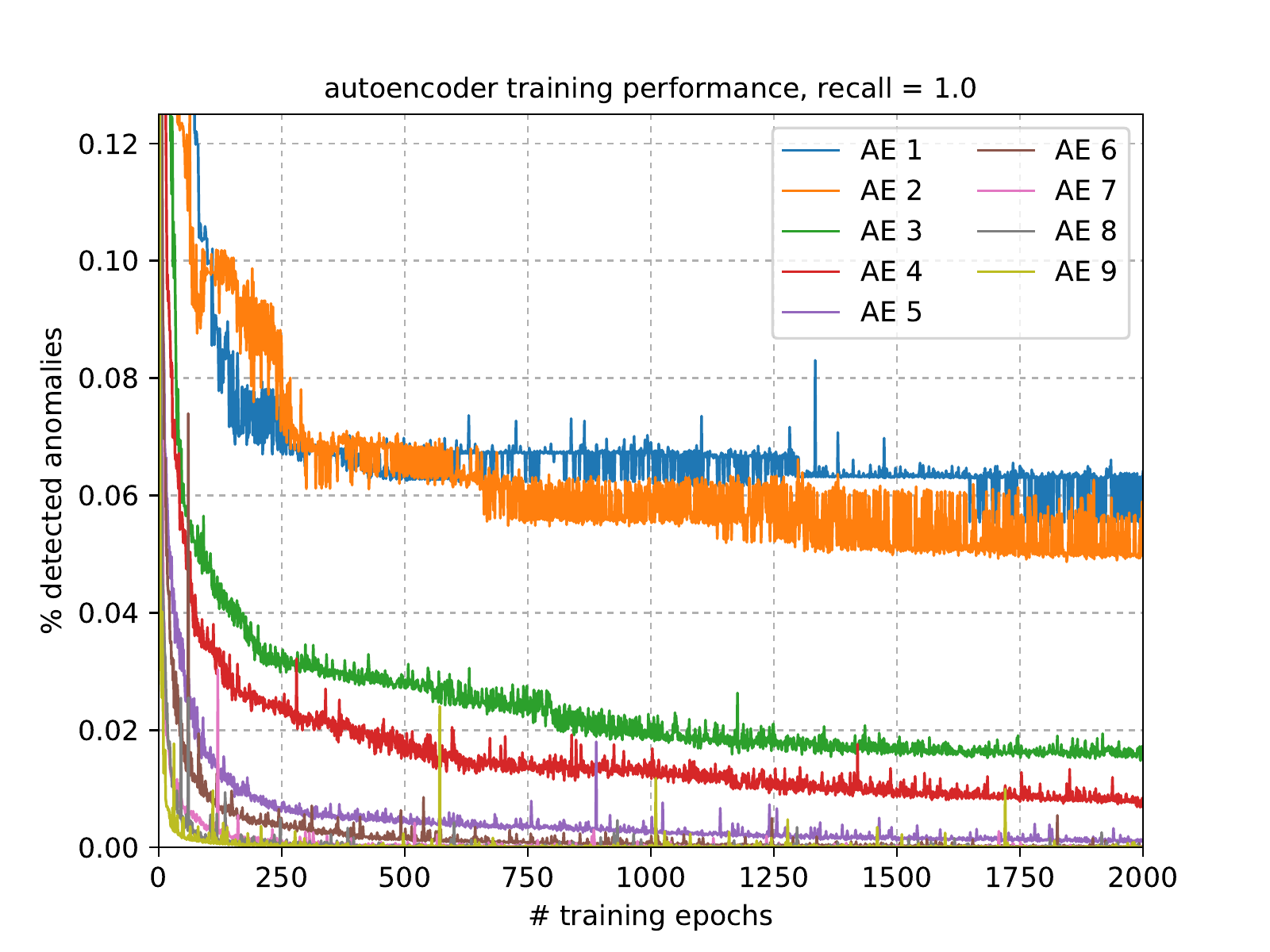}
\endminipage\hfill
\minipage{0.5\textwidth}
	\includegraphics[width=\linewidth]{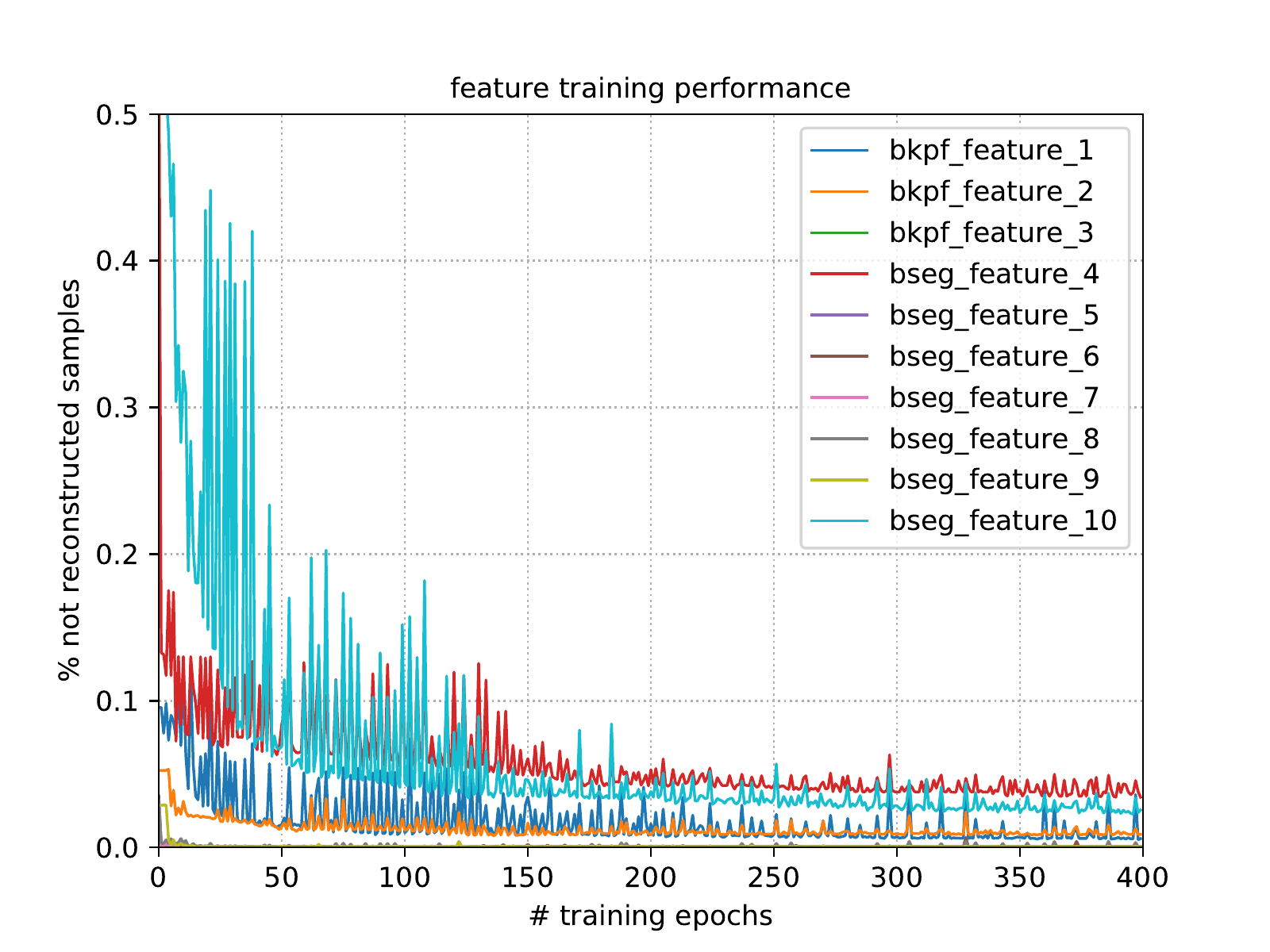}
\endminipage\hfill
	\caption{Training performance using dataset A of the evaluated autoencoder architectures AE 1 - AE 9 (left). Training performance using dataset B of the individual journal entry attributes (right).}
    \label{fig:training_progress}
\end{figure}

In annual audits auditors aim to limit the number of journal entries subject to substantive testing to not miss any error or fraud related entry. Derived from this desire we formulated three objectives guiding our training procedure: (1) minimize the overall autoencoder reconstruction error, (2) focus on models that exhibit a recall of 100\% of the synthetic journal entries, and (3) maximize the autoencoder detection precision to reduce the number of false-positive alerts.

We trained nine distinct architectures ranging from shallow (AE 1) to deep (AE 9) autoencoder networks. Table \ref{fig:topologytable} shows an overview of the evaluated architectures \footnote{The notation: $[401; 576]-3-[401; 576]$, denotes a network architecture consisting of three fully connected layers. An input-layer consisting of 401 or 576 neurons (depending on the encoded dimensionality of the dataset respectively), a hidden layer consisting of 3 neurons, as well as, an output layer consisting of 401 or 576 neurons.} The depth of the evaluated architectures was increased by continuously adding fully-connected hidden layers of size $2^k$ neurons, where $k=2,3,...,9$. To prevent saturation of the non-linearities we choose leaky rectified linear units (LReLU) \cite{Xu2015} and set their scaling factor to $a=0.4$. 


Each autoencoder architecture was trained by applying an equal learning rate of $\eta = 10^{-4}$ to all layers and using a mini-batch size of $128$ journal entries. Furthermore, we used adaptive moment estimation \cite{Kingma2015} and initialized the weights of each network layer as proposed in \cite{Glorot2010}. The training was conducted via standard back-propagation until convergence (max. 2'000 training epochs). For each architecture, we run the experiments five times using distinct parameter initialization seeds to guarantee a deterministic range of result.

Figure \ref{fig:training_progress} (left) illustrates the performance of the distinct network topologies evaluated for dataset A over progressing training epochs. Increasing the number of hidden units results in faster error convergence and decreases the number of detected anomalies. Figure \ref{fig:training_progress} (right) shows the individual attribute training performance of dataset B with progressing training. We noticed that the training performance of individual attributes correlates with the number of distinct attributes values e.g. the attribute $bseg\_attrubute\_10$ exhibits a total of 19 distinct values whereas $bseg\_attribute\_2$ exhibits only 2 values and is therefore learned faster. 

\begin{figure}[t]
\minipage{1.00\textwidth}
	\includegraphics[scale=0.55, angle=0, trim= 38 0 0 0]{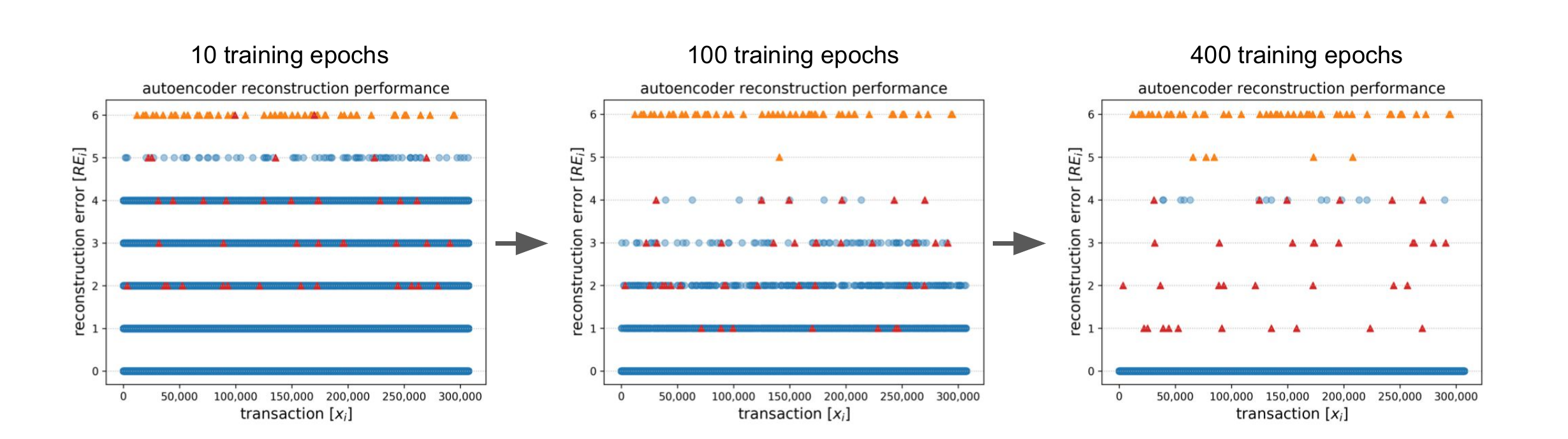}
\endminipage
	\caption{Journal entry reconstruction error $RE$ obtained for each of the 307.457 journal entries $x_i$ contained in dataset A after 10 (left), 100 (middle) and 400 (right) training epochs. The deep autoencoder (AE 8) learns to distinguish global anomalies (orange) and local anomalies (red) from original journal entries (blue) with progressing training epochs.}
	\label{fig:training_results}
\end{figure}

Once the training converged the trained models are used to obtain the reconstruction errors $RE$ of each journal entry. Figure \ref{fig:training_results} illustrates the reconstruction errors obtained for the 307'457 journal entries contained in dataset A after 10 (left), 100 (middle) and 400 (right) training epochs. The used autoencoder (AE 8) learns to reconstruct the majority of original journal entries (blue) with progressing training epochs and fails to do so for the \textit{global anomalies} (orange) and \textit{local anomalies} (red).

We set the anomaly threshold $\beta = 0.01$ implying that a journal entry is labeled "anomalous" if one of its attributes was not reconstructed correctly or occurs very rarely. This was done in compliance with real-world audit scenarios in which auditors tend to handle fraudulent journal entries in a conservative manner to mitigate risks and not miss a potential true positive.

\section{Experimental Results}
\label{sec:results}

This section describes the results of our evaluation. Upon successful training we evaluated the proposed scoring according to two criteria: (1) Are the trained autoencoder architectures capable of learning a model of the regular journal entries and thereby detect the injected anomalies ("quantitative" evaluation)? (2) Are the detected and non-injected anomalies "suspicious" enough to be followed up by accountants or forensic examiners ("qualitative" evaluation)?

 \begin{figure}[t]
 \minipage{0.5\textwidth}
	\includegraphics[width=\linewidth]{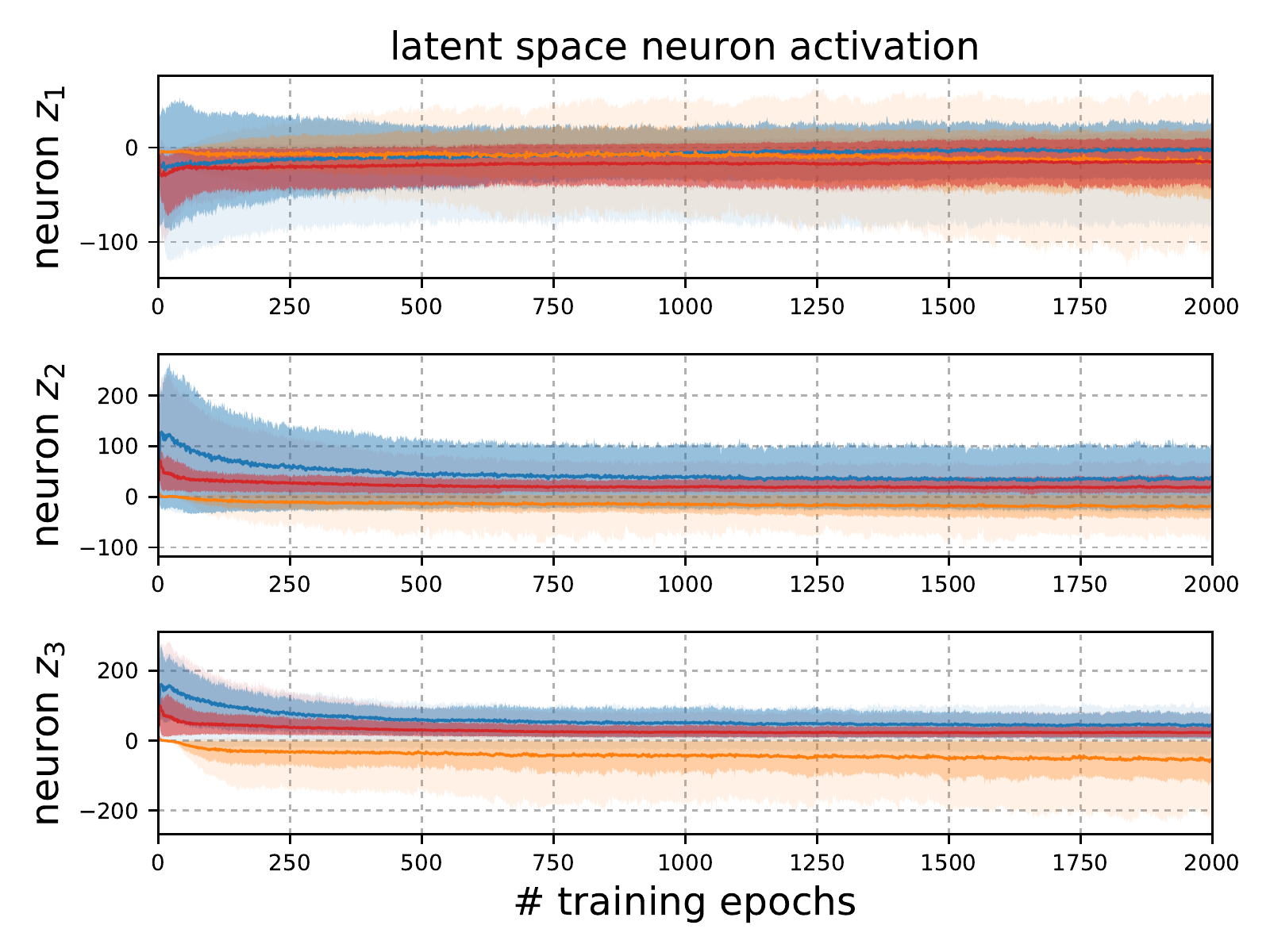}
\endminipage\hfill
\minipage{0.5\textwidth}
	\includegraphics[width=\linewidth]{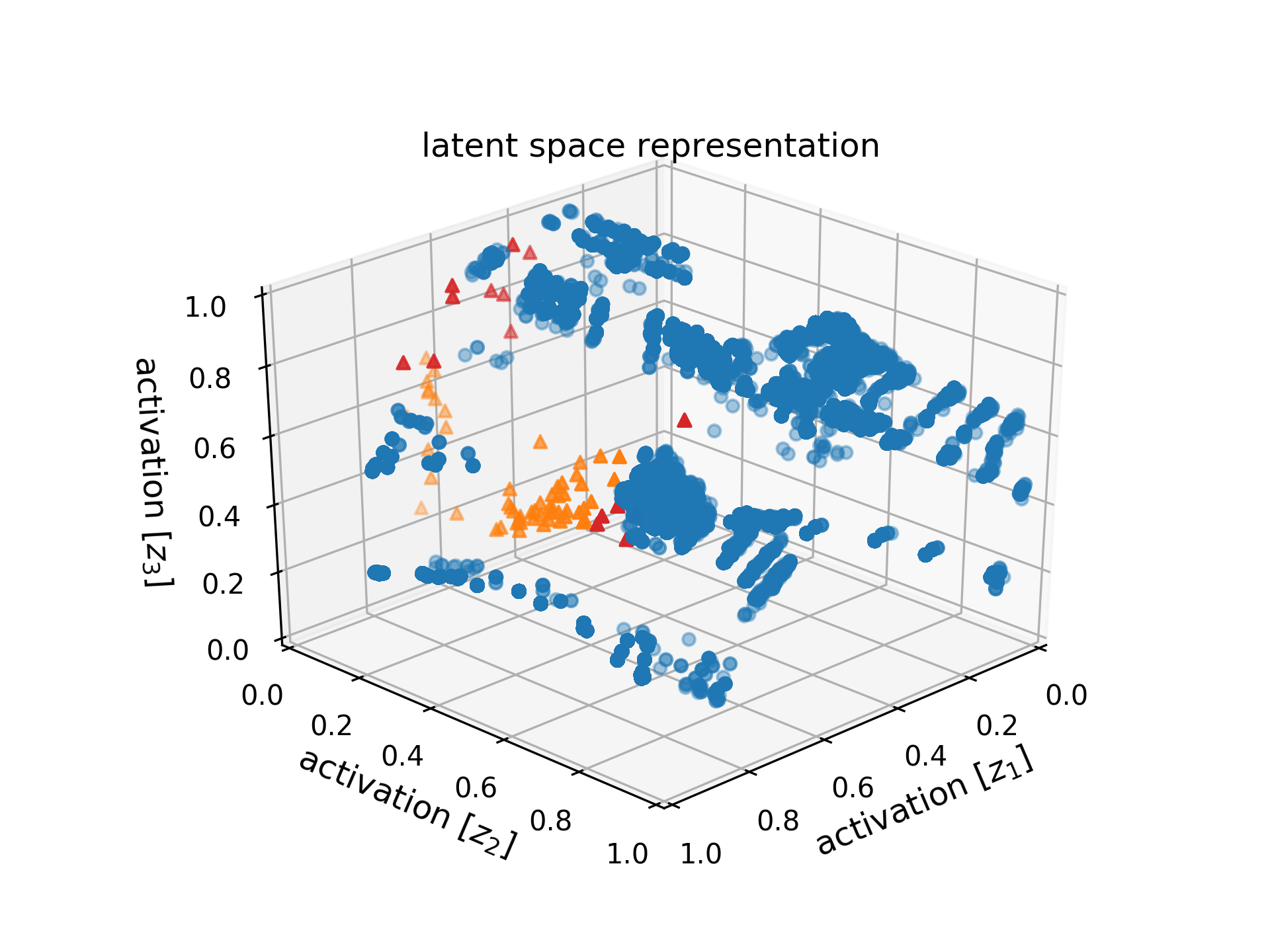}
\endminipage\hfill
	\caption{Learned latent space neuron activations $z_1$, $z_2$ and $z_3$ of autoencoder architecture AE 9 and dataset A with progressing training epochs (left). Latent space representation of dataset A learned by autoencoder architecture AE 9 after 2'000 training epochs (left). }
    \label{fig:latent_space}
\end{figure}

\begin{table}[t]
  \fontsize{8}{6}\selectfont
  \centering
  \begin{tabular}{c c c c c c c c}
   \toprule
    \multirow{2}{*}{Model}&\multirow{2}{*}{Dataset}&\multirow{2}{*}{Precision}&\multirow{2}{*}{F\textsubscript{1}-Score}&\multirow{2}{*}{Top-k}&\multicolumn{1}{c}{Anomalies}&\multicolumn{1}{c}{Anomalies}\\ &&&&&\multicolumn{1}{c}{[\%]}&\multicolumn{1}{c}{[\#]}\\
    \midrule
    AE 1&A&0.0049&0.0098&0.0049&6.26&\multicolumn{1}{r}{19'233}\hspace{0.2cm}\\
    AE 2&A&0.0063&0.0126&0.0063&4.87&\multicolumn{1}{r}{14'966}\hspace{0.2cm}\\
    AE 3&A&0.0098&0.0194&0.6632&3.16&\multicolumn{1}{r}{9'719}\hspace{0.2cm}\\
    AE 4&A&0.0290&0.0564&0.7684&1.07&\multicolumn{1}{r}{3'275}\hspace{0.2cm}\\
    AE 5&A&0.0641&0.1204&0.6632&0.48&\multicolumn{1}{r}{1'483}\hspace{0.2cm}\\
    AE 6&A&0.0752&0.1398&0.5263&0.41&\multicolumn{1}{r}{1'264}\hspace{0.2cm}\\
    AE 7&A&0.0796&0.1474&0.7895&0.39&\multicolumn{1}{r}{1'194}\hspace{0.2cm}\\
    AE 8&A&0.1201&0.2144&0.5684&0.26&\multicolumn{1}{r}{791}\hspace{0.2cm}\\
    \textbf{AE 9}&\textbf{A}&\textbf{0.1971}&\textbf{0.3293}&\textbf{0.6947}&\textbf{0.16}&\multicolumn{1}{r}{\textbf{482}}\hspace{0.2cm}\\
    \midrule
    AE 1&B&0.0020&0.0040&0.0020&28.84&\multicolumn{1}{r}{49'897}\hspace{0.2cm}\\
    AE 2&B&0.0030&0.0059&0.0030&19.52&\multicolumn{1}{r}{33'762}\hspace{0.2cm}\\
	AE 3&B&0.0052&0.0104&0.6200&11.04&\multicolumn{1}{r}{19'102}\hspace{0.2cm}\\
    AE 4&B&0.0076&0.0150&0.7300&7.65&\multicolumn{1}{r}{13'238}\hspace{0.2cm}\\
    AE 5&B&0.0087&0.0173&0.7400&6.62&\multicolumn{1}{r}{11'444}\hspace{0.2cm}\\
    AE 6&B&0.0251&0.0489&0.6100&2.30&\multicolumn{1}{r}{3'986}\hspace{0.2cm}\\
    AE 7&B&0.0268&0.0522&0.6400&2.15&\multicolumn{1}{r}{3'735}\hspace{0.2cm}\\
    AE 8&B&0.0197&0.0387&0.6700&2.93&\multicolumn{1}{r}{5'070}\hspace{0.2cm}\\
    \textbf{AE 9}&\textbf{B}&\textbf{0.0926}&\textbf{0.1695}&\textbf{0.4200}&\textbf{0.62}&\multicolumn{1}{r}{\textbf{1'080}}\hspace{0.2cm}\\
    \bottomrule \\   
  \end{tabular}
 \caption{Evaluation of anomaly detection performance for both datasets A and B using autoencoder architectures AE 1 - AE 9. The training was constrained to models exhibiting a recall of 100\% of the synthetic journal entries using standard back propagation. Models were trained until reconstruction error convergence or max. 2'000 epochs. The best detection performances was obtained by architecture AE 9 consisting of 17 hidden layers and LReLU activations.}
   \label{fig:resulttableA}
 \end{table}
 
\subsection{Quantitative Evaluation} To quantitatively evaluate the effectiveness of the proposed approach, a range of evaluation metrics including precision, standard f\textsubscript{1}-Score, top-k precision, absolute and relative number of detected anomalies are reported. The choice of f\textsubscript{1}-Score is to account for the highly unbalanced anomalous vs. non-anomalous class distribution of the datasets. To calculate the top-k precision we set $k=95$ (dataset A) and $k=100$ (dataset B) corresponding to the number of synthetic anomalies in both benchmark datasets.



\begin{figure}[t]
\minipage{0.5\textwidth}
	\includegraphics[width=\linewidth]{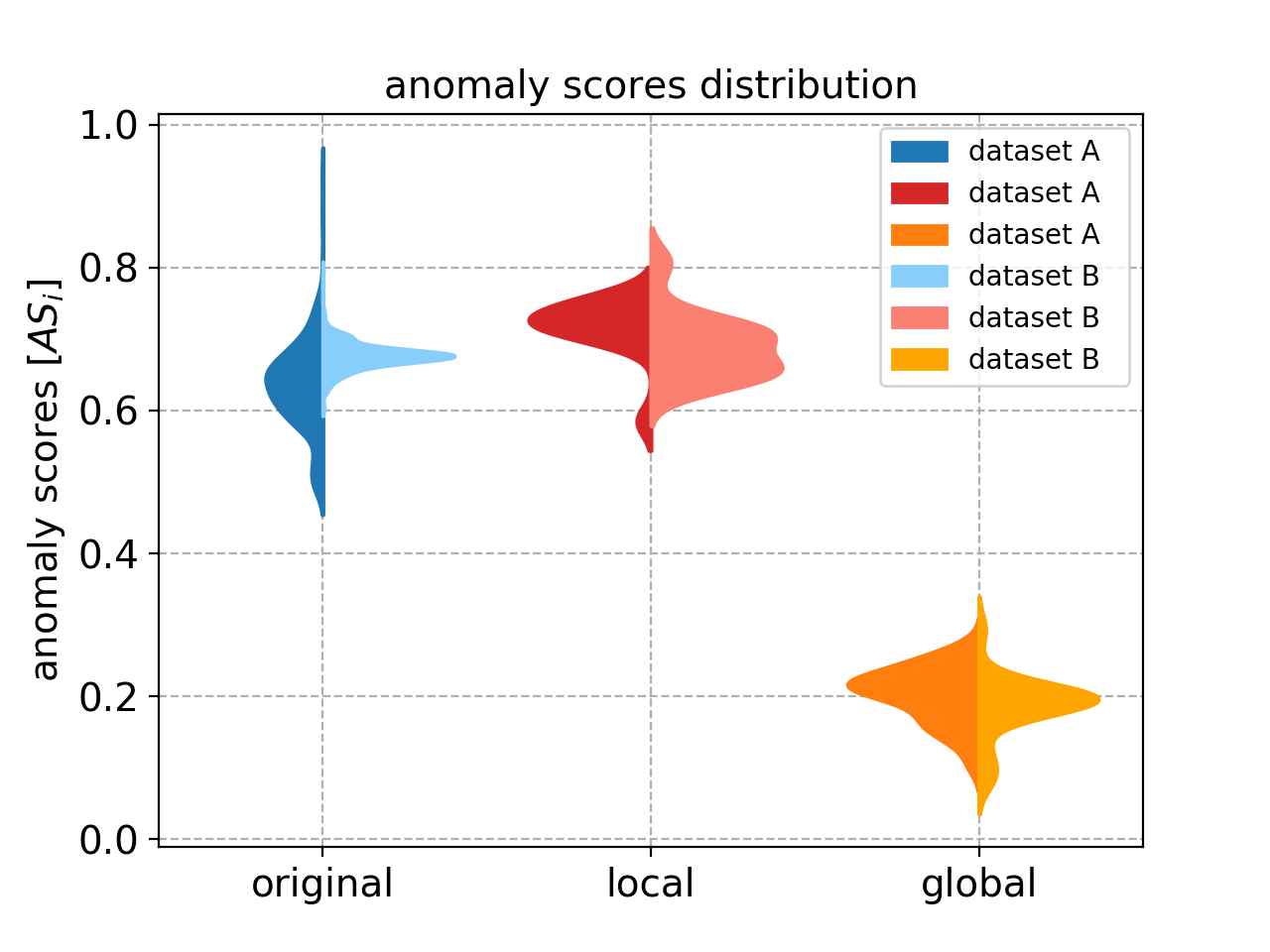}
\endminipage\hfill
\minipage{0.5\textwidth}
	\includegraphics[width=\linewidth]{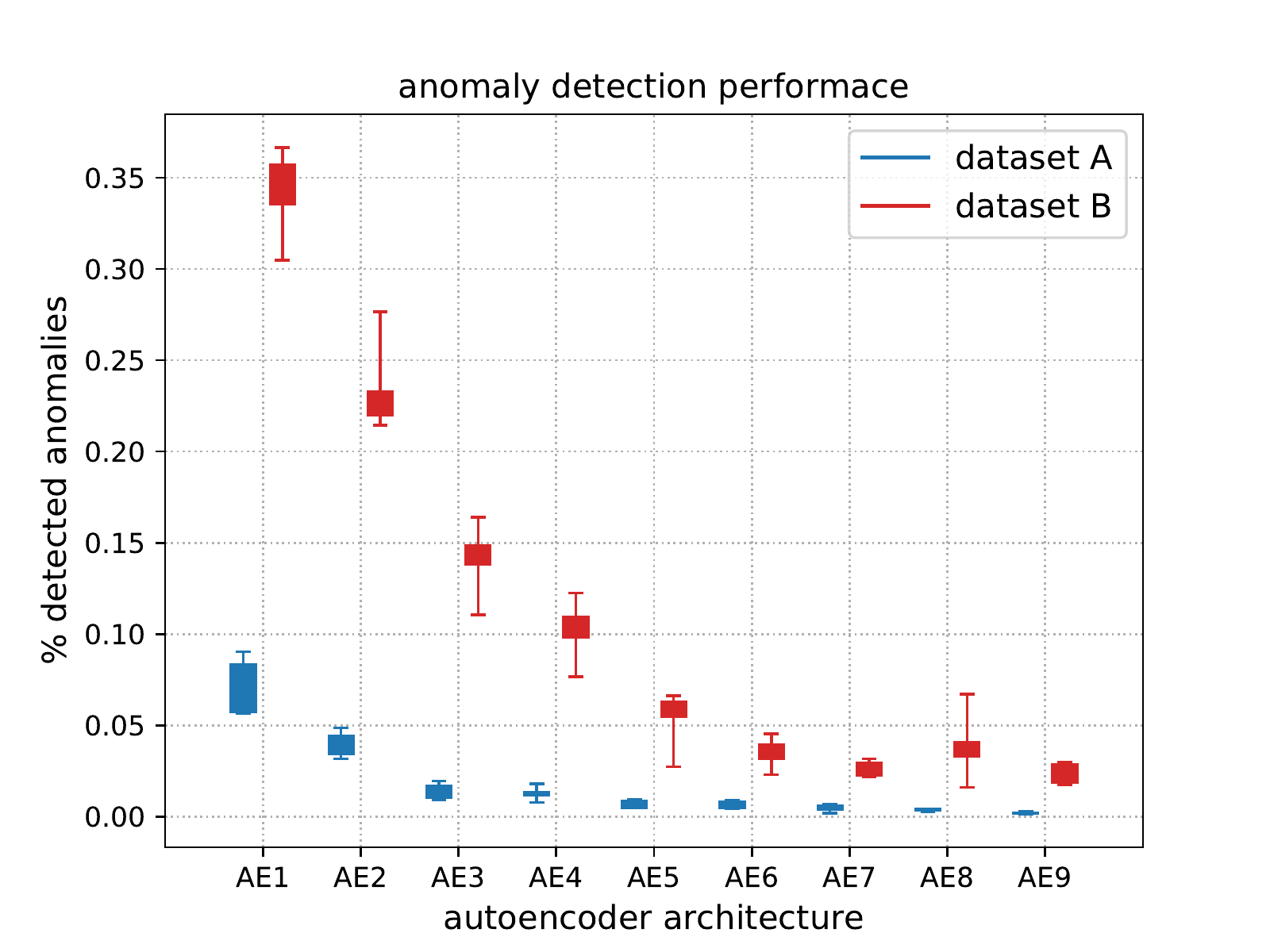}
\endminipage\hfill
\caption{Distribution of anomaly scores $AS_i$ of the distinct journal entry classes in both datasets using a trained deep autoencoder (AE 9) and $\alpha=0.3$ (left). Fraction of detected anomalies in both datasets for the evaluated autoencoder architectures (right). Result variation originates from initializing the weights of each model with five distinct seed values.}
    \label{fig:roc_results}
\end{figure}

Table \ref{fig:resulttableA} shows the obtained results of both benchmark datasets using distinct network architectures. Increasing the number of hidden layers reduces the number of detected anomalies. While preserving recall of 100\%, the deepest trained autoencoder architecture (AE 9) results in a low fraction of 0.16\% detected anomalies in dataset A and 0.62\% detected anomalies in dataset B. The observed results show that the autoencoder depth substantially affects its ability to model the inherent manifold structure within each dataset. Figure \ref{fig:roc_results} (left) illustrates the anomaly score $AS_i$ distributions of the distinct journal entry classes using a trained deep autoencoder (AE 9) and $\alpha = 0.3$. 

To understand the observed difference in detection performance for both datasets we also investigated the learned latent space representations. Figure \ref{fig:latent_space} (left) shows the mean neuron activation $[z_1, z_2, z_3]$ of the deep autoencoder (AE 9) three "bottleneck" neurons $z_1$, $z_2$ and $z_3$. With progressing training epochs the network learns a distinctive activation pattern for each journal entry class. Upon 2'000 training epochs a mean activation of $[-12.99, -20.57, -50.55]$ can be observed for çtextit{global anomalies} (orange), $[-15.09, 18.99, 23.40]$ for \textit{local anomalies} (red) and $[-1.67, 37.28, 44.76]$ for regular journal entries (blue). Figure \ref{fig:latent_space} (right) shows the learned manifolds structure after the training completion.



\subsection{Qualitative Evaluation} To qualitatively evaluate the character of the detected anomalies contained in both datasets we reviewed all non-synthetic journal entries detected by the AE 9 architecture. To distinguish local from global anomalies we empirically choose to set $\alpha= 0.3$ and flagged the journal entries exhibiting an $AS \geq 0.4$ as local anomalies and $AS < 0.4$ as global anomalies. 
 
As anticipated the review of the \textit{global anomalies} revealed that the majority of anomalies correspond to journal entries that exhibit one or two rare attribute values e.g. journal entries that correspond to seldom vendors or seldom currencies. In addition, we also detected journal entries referring to: (1) posting errors due to wrongly used general ledger accounts; (2) journal entries of unusual document types containing extremely infrequent tax codes; and (3) incomplete journal entries exhibiting missing currency information. Especially, the latter observations indicated a weak control environment around certain business processes of the investigated organization. 
 
The review of the \textit{local anomaly} journal entry population showed that these anomalies correspond to journal entries exhibiting attributes that are frequently observable but rarely occur in combination e.g. changes of business process or rarely applied accounting practices. A more detailed investigation of the detected instances uncovered: (1) shipments to customers that are invoiced in different than the usual currency; (2) products send to a regular client but were surprisingly booked to another company code; (3) postings that exhibit an unusual large time lag between document date and posting date; and, (4) irregular rental payments that slightly deviate from ordinary payments. Our initial feedback received by auditors underpinned not only their relevance from an audit but also a forensic perspective.

 \begin{table}[t]
  \fontsize{8}{6}\selectfont
  \centering
  \begin{tabular}{l c c c c c r}
    \toprule
        \multirow{2}{*}{Method}&\multirow{2}{*}{Data}&\multirow{2}{*}{Precision}&\multirow{2}{*}{F\textsubscript{1}-Score}&\multicolumn{1}{c}{ROC}&\multicolumn{1}{c}{Anomalies}&\multicolumn{1}{c}{Anomalies}\\
        &&&&\multicolumn{1}{c}{AUC}&\multicolumn{1}{c}{[\%]}&\multicolumn{1}{c}{[\#]}\\
    \midrule
    PCA {\tiny(c=30)} &A&0.0055&0.0110&0.9608&1.63&17'162\\ 
    HDBSCAN {\tiny(mcl=1'100)} &A&0.0256&0.0499&0.9787&1.29&3'714\\ 
    LOF {\tiny(k-NN=50)} &A&0.0499&0.0952&0.9979&0.62&1'901\\ 
    OC-SVM {\tiny(nu=0.005, $\gamma$=0.95)} &A&0.2769&0.4338&0.9999&0.11&343\\ 
    \textbf{AE {\tiny(m=AE 9)}}&A&\textbf{0.5688}&\textbf{0.7251}&\textbf{0.9999}&\textbf{0.05}&\textbf{167}\\
    \midrule
    PCA {\tiny(c=36)}&B&0.0331&0.0640&0.9978&1.75&3'025\\
    HDBSCAN {\tiny(mcl=7'445)}&B&0.0383&0.0720&0.9744&1.65&2'680\\
    LOF {\tiny(k-NN=10)} &B&0.0518&0.0986&0.9984&1.12&1'929\\
    OC-SVM {\tiny(nu=0.01, $\gamma$=0.95)} &B&0.1397&0.2451&0.9997&0.41&716\\
    \textbf{AE {\tiny(m=AE 9)}}&B&\textbf{0.1616}&\textbf{0.2782}&\textbf{0.9997}&\textbf{0.35}&\textbf{619}\\
    \bottomrule \\
  \end{tabular}
 \caption{Comparative evaluation of the autoencoder based approach against several unsupervised and non-parameteric anomaly detection techniques (best performing parameters of each technique in brackets). For each method the best detection performance is reported that results (1) in a recall of 100\% of the synthetic anomalies and (2) the best obtainable ROC-AUC.}
    \label{fig:resulttableB}
 \end{table}
 
\subsection{Baseline Evaluation:} We evaluated the autoencoder network based approach against unsupervised and non-parametric anomaly detection techniques, namely (1) reconstruction error-based: Principal Component Analysis (PCA) \cite{Pearson1901}, (2) kernel-based: One Class Support Vector Machine (OC-SVM) \cite{Scholkopf1999}, (3) density based: Local-Outlier Factor (LOF) \cite{Breunig2000}, and (4) hierarchical nearest-neighbor based: Density-Based Spatial Clustering of Applications with Noise (DBSCAN) \cite{Ester1996} anomaly detection. For all methods, besides DBSCAN, the performance was assessed using their implementations of the sci-kit machine learning library \cite{Pedregosa2011}. For DBSCAN we used the optimized HDBSCAN implementation developed by Campello et al.\cite{Rahman2016}. 

To conduct a fair comparison an exhaustive grid search over each techniques parameter space was conducted to determine their best performing parameters. We report the anomaly detection performance of each technique in table \ref{fig:resulttableB}. The best performing results are selected based on parameterizations that (1) result in a recall of 100\% of the synthetic anomalies and correspond to (2) the highest area under the ROC curve (ROC-AUC). 


For both evaluation datasets, best ROC-AUC results are obtained for the OC-SVM and the AE 9 autoencoder architecture. However, the autoencoder based approach outperforms the other benchmark techniques in terms of its anomaly detection precision. Comparing both the OC-SVM and the AE 9 for dataset A, the autoencoder results in 176 less detected false positive anomalies; while for dataset B, the autoencoder results in 97 less detected false positive anomalies. The lower but highly accurate number of false positive alerts is of great relevance in the context of real financial audit scenarios where substantiative evaluation of a single detected anomaly can results in considerable effort. 

\section{Conclusion and Future Work}
\label{sec:conclusion}


In this work we presented the first deep learning based approach for the detection of anomalous journal entries in large scaled accounting data. Our empirical evaluation using two real-world accounting datasets demonstrates that the reconstruction error of deep autoencoder networks can be used as a highly adaptive anomaly assessment of journal entries. In our experiments we achieved a superior f\textsubscript{1}-score of 32.93 in dataset A and 16.95 in dataset B compared to state of the art baseline methods. Qualitative feedback, received by auditors and forensic accountants, on the detected anomalies underpinned that our method captures journal entries of high relevance for a detailed follow-up audit. 

We are excited about the future of deep learning based audit approaches and plan to conduct a more detailed investigation of the journal entries' latent space representations learned by deep autoencoders. We believe that investigating the latent manifolds will provide additional insights into the fundamental structures of accounting data and underlying business processes. Furthermore, we aim to evaluate the anomaly detection ability of more recently proposed autoencoder architectures e.g. adversarial autoencoder neural networks. 

Given the tremendous amount of journal entries recorded by organizations annually, an automated and high precisions detection of accounting anomalies can save auditors considerable time and decrease the risk of fraudulent financial statements. \\

\noindent The code we used to train and evaluate our models is available at: \\ https://github.com/GitiHubi/deepAI.

 
\subsubsection*{Acknowledgments}
This work was supported by the NVIDIA AI Lab (NVAIL) program. We thank Adrian Ulges, all members of the Deep Learning Competence Center at the DFKI, as well as, the PwC Europe's Forensic Services team for their comments and support.

%
%
%

\bibliographystyle{splncs04}
\bibliography{bibtex/library}

%




\end{document}